\documentclass[runningheads]{llncs}

\usepackage{eccv}

\usepackage{siunitx}
\usepackage{tcolorbox}
\usepackage{multirow}
\usepackage{colortbl}

\usepackage{placeins}

\usepackage{eccvabbrv}

\usepackage{graphicx}
\usepackage{booktabs}
\usepackage{capt-of}
\usepackage{comment}

\usepackage[accsupp]{axessibility}  

\usepackage{hyperref}

\usepackage{orcidlink}

\begin{document}

\title{Zero-Shot 3D Plant Organ Segmentation with SAM3 and Semantic NeRFs} 


\author{Andreas Gilson\inst{1,2}\orcidlink{0009-0001-1674-0447} \and
Laura Hennig\inst{1,3} \and
Peter Pietrzyk\inst{1}\orcidlink{0000-0002-6794-8133}}

\authorrunning{A.~Gilson et al.}

\institute{Fraunhofer Institute for Integrated Circuits (IIS), Fürth, Germany \and
Otto-Friedrich-Universität Bamberg, Germany \and Friedrich-Alexander-Universität Erlangen-Nürnberg, Germany}

\maketitle

\begin{abstract}
Accurate 3D plant organ segmentation is fundamental to automated phenotyping. Existing approaches rely on annotated training data or species-specific model configurations. We present an annotation-free pipeline for 3D plant organ segmentation, combining text-prompted SAM3 segmentation with semantic neural radiance fields (NeRFs). Given only multi-view RGB images and a list of class names, our zero-shot pipeline produces semantically labeled 3D point clouds without manual annotation, per-species fine-tuning, or domain-specific preprocessing. Multi-view NeRF fusion acts as effective implicit consensus mechanism that lifts imperfect per-frame masks into accurate 3D labels. On a controlled \textit{Begonia maculata} testbed the SAM3 pipeline achieves 92.6\% mIoU, reaching 95.9\% of the oracle upper bound established with perfect ground-truth masks. The pipeline was further evaluated on a new dataset spanning ten diverse plant point clouds reaching an average $0.856\pm0.097$ mIoU, with leaf and pot IoU above 0.91 and 0.90 for every species, respectively. These results demonstrate that annotation-free 3D plant organ segmentation is now feasible and approaching the range of supervised methods.

\end{abstract}

\section{Introduction}
\label{sec:intro}

Reliable 3D point cloud segmentation is a key prerequisite for 3D plant phenotyping to enable accurate trait extraction~\cite{chang_ai-driven_2026}. Most existing methods segment 3D point clouds using supervised deep learning methods, which rely on labeled training data~\cite{xie_delving_2024}. This is a bottleneck for plant organ segmentation, since annotating data in 3D is labor- and time-intensive work ~\cite{ghose_advancements_2026}. This gap is slowly closing with an increasing number of annotated plant datasets being published. For setups with multi-view imaging sensors, it is also possible to utilize 2D detectors and project the 2D annotations to 3D~\cite{yang_plantsegnerf_2026,chen_teanerf_2026}. Furthermore, research in 2D foundation models for image segmentation is advancing rapidly, with SAM3~\cite{carion_sam_2025} being the most recent. While the idea of using 2D foundation models for 3D plant segmentation has been investigated already for fruit detection~\cite{meyer_fruitnerf_2024}, we are not aware of any work performing open-vocabulary 3D plant organ segmentation completely without requiring manual annotations. By combining SAM3 with semantic NeRFs, we propose a pipeline for \textbf{plant point cloud segmentation that works zero-shot without any labels}. Based on simple text prompts our pipeline achieves strong results for plant segmentation. This paper makes the following contributions:

\begin{enumerate}
    \item We present a pipeline that lifts text-prompted SAM3 segmentation into 3D via a neural radiance field, requiring zero annotated training images.
    \item We demonstrate and evaluate this on a novel dataset for 3D plant segmentation for ten different species.
    \item We identify SAM3's bimodal confidence distribution as a key property that simplifies pipeline deployment and makes multi-view fusion an effective consensus mechanism.
\end{enumerate}

\section{Related Work}
\label{sec:related_work}

Deep learning on 3D point clouds has become the dominant paradigm for plant organ segmentation, with general-purpose backbones such as PointNet\cite{qi_pointnet_2017}, PointNet++\cite{qi2017pointnetplusplus}, and DGCNN\cite{wang_dynamic_2019} underpinning most current methods. However, plant phenotyping introduces domain-specific challenges: high morphological variability, thin structures, and severe class imbalance\cite{kamal_systematic_2026}. Several researchers address these challenges with plant-specific adaptions \cite{mertoglu_planest-3d_2024,wang_vgds-pointnet_2026,ma_tsinet_2025}. The specialized architecture of PlantNet's\cite{li_plantnet_2022} successor PSegNet\cite{li_psegnet_2022} combines edge-preserving down-sampling with semantic- and instance-feature fusion, achieving up to 89.90\% mIoU across tobacco, tomato, and sorghum. OmniPlantSeg\cite{gilson_omniplantseg_2025} avoids down-sampling and Organ3DNet\cite{li_organ3dnet_2026} overcomes the fixed-size limitation with a Transformer based architecture, reaching state-of-the-art performance (93.57\% IoU, 94.69\% mWCov) on a five-species dataset.

All supervised methods share one common limitation: they require dense point-wise annotations. 
This, together with the progress made by vision foundation models motivated a new line of methods that instead obtain 3D labels from 2D image supervision.

An alternative to direct 3D annotation is to predict labels in 2D and lift them into 3D representations\cite{zhi_-place_2021}. This is done using radiance field technology like NeRFs\cite{mildenhall_nerf_2020} and 3D Gaussian Splatting (3DGS)\cite{kerbl_3d_2023}. The expansion with a semantic output head allows per-pixel 2D predictions to be fused into view-consistent 3D models via differentiable rendering. General-scene works have demonstrated this for closed-vocabulary semantics\cite{zhi_-place_2021}, open-vocabulary queries\cite{kerr_lerf_2023}, interactive segmentation from sparse prompts\cite{cen_segment_2023}, and instance-level lifting via contrastive features on both NeRFs\cite{bhalgat_contrastive_2023, siddiqui_panoptic_2022} and Gaussians\cite{ye_gaussian_2024,lyu_gaga_2024}. In plant phenotyping, FruitNeRF\cite{meyer_fruitnerf_2024} and FruitNeRF++\cite{meyer_fruitnerf_2025} lift binary and instance-level masks for fruits into NeRFs for counting. TeaNeRF\cite{chen_teanerf_2026} and PlantSegNeRF\cite{yang_plantsegnerf_2026} combine trained detectors (YOLOv11) with semantic or instance NeRF branches for organ-level segmentation. Several recent studies have also combined SAM-based image segmentation with Gaussian Splatting for crop reconstruction \cite{yang_soybeaninsgs_2026,jiang_cotton3dgaussians_2025,shen_biomass_2025}. LeafFit\cite{luo_leaffit_2026} performs training-free geodesic leaf segmentation directly on Gaussian primitives. Despite reducing annotation from 3D to 2D, all plant-specific methods still require either a trained per-species 2D detector\cite{yang_plantsegnerf_2026, chen_teanerf_2026, yang_soybeaninsgs_2026} or single target classes \cite{meyer_fruitnerf_2024}. Cross-view instance consistency must be explicitly enforced through contrastive losses, voting schemes, or tracking algorithms. 

Until recently, foundation models for segmentation were not powerful enough to replace trained per-species detectors. This changed with Grounded-SAM\cite{ren_grounded_2024} that couples the Segment Anything Model (SAM) \cite{kirillov_segment_2023} with an open-vocabulary detector (Grounding DINO\cite{liu_grounding_2024}): text prompts are first converted to bounding boxes, which then serve as SAM's spatial input. This two-model pipeline enabled text-driven 2D segmentation but introduced additional complexity and error propagation between detection and segmentation stages. SAM2 \cite{ravi_sam_2024} improved temporal consistency for video inputs but retained the same fundamental limitation: it segments \textit{at} a given location rather than \textit{finding} a given concept. SAM3\cite{carion_sam_2025} overcomes this by introducing promptable concept segmentation as a new paradigm\cite{sapkota_sam2--sam3_2025}. SAM3 fuses image features with text prompt embeddings via cross-attention, enabling native text-conditioned instance segmentation without an external detector. For 3D plant segmentation, SAM3's capabilities open new research opportunities: rather than training a YOLO or fine-tuning Grounding DINO per species, a single text prompt (e.g, \textit{leaf}, \textit{stem}) suffices to generate 2D masks for multi-view image sets that can be used for lifting into 3D. We combine the zero-shot capability of SAM3 with neural radiance fields to achieve 3D plant segmentation without any domain-specific annotation or model training.

\section{Method}
\subsection{Overview}

Our pipeline consists of five steps. 2D images are acquired from any source (i) are used by Colmap for camera pose estimation via structure-from-motion (ii) and simultaneously segmented by SAM3 with pre-defined text prompts (iii). The results are used to train a Nerfacto model with semantic head (iv). Finally, the point cloud is exported from the NeRFs color-, density- and semantic-fields. Our research hypothesis for the experiments is:

\begin{figure}[htbp]
    \centering
    \includegraphics[width=1\linewidth]{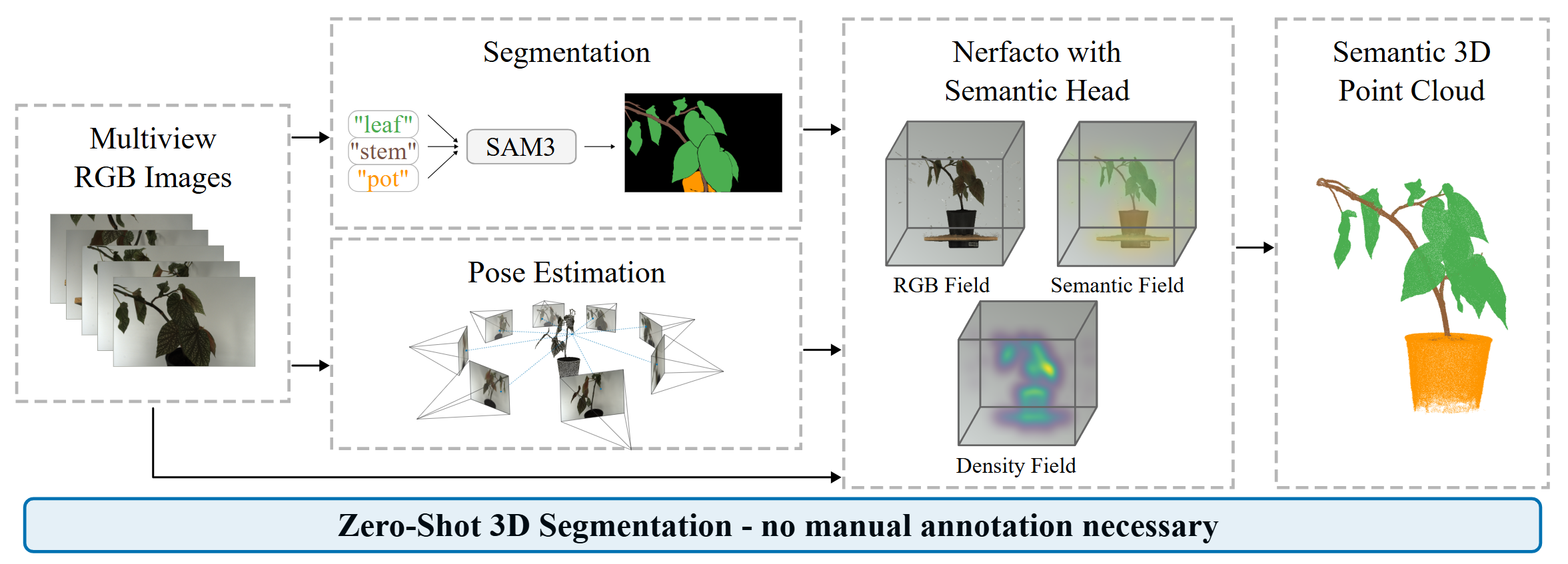}
    \caption{Overview of the zero-shot pipeline. Multi-view RGB images are segmented by text-prompted SAM3 into semantic masks. Images, masks and poses are used to train multi-class semantic NeRF that exports semantic 3D point cloud.}
    \label{fig:overview}
\end{figure}

\noindent\textbf{Research hypothesis:}\quad
\textit{Multi-view fusion acts as an implicit consensus mechanism. Since each 3D point is supervised from multiple viewpoints, correct labels from the majority of views outweigh occasional label errors without requiring any explicit consistency enforcement, voting scheme, or contrastive loss.}

\subsection{Text-Prompted 2D Segmentation (SAM3)}
\label{sec:sam3}

We generate per-image semantic labels using SAM3\cite{carion_sam_2025}, a vision-language foundation model that unifies object detection and segmentation through text-conditioned mask generation. SAM3 combines a 450M-parameter ViT-based vision encoder with a 300M-parameter causal text encoder trained on 5.4B image--text pairs via contrastive learning and a DETR-style transformer decoder with 200 learned object queries\cite{carion_sam_2025}.
Unlike SAM\cite{kirillov_segment_2023} and SAM2\cite{ravi_sam_2024}, which require spatial prompts (points, boxes, or masks) to indicate \emph{where} to segment, SAM3 accepts natural-language prompts that describe \emph{what} to segment.

We prompt SAM3 in FP32 precision with the plain class names such as \texttt{"leaf"}, \texttt{"stem"}, and \texttt{"pot"}.

Each image is processed independently and no cross-view linking or instance tracking is performed at this stage. For each prompt, SAM3 returns zero or more instance masks with associated confidence scores. Predictions below a confidence threshold $\tau = 0.5$ are discarded.

To produce a single semantic label map per image, prompts are processed in priority order with higher-priority classes overwriting lower-priority ones at overlapping pixels (priority: leaf $>$ stem $>$ pot). All instance masks of the same class are merged into one semantic class. Pixels not claimed by any foreground prompt receive the background label (class~0) and are excluded from the downstream semantic loss via class weighting. The resulting output per image is a class-index map with $C = |\text{prompts}| + 1$ classes, including the background.

\subsection{Semantic NeRF Training}
\label{sec:nerf}

Our 3D pipeline builds on FruitNeRF~\cite{meyer_fruitnerf_2024}, a semantic extension of Nerfacto implemented in nerfstudio~\cite{tancik_nerfstudio_2023}, which we adapt for multi-class organ segmentation. The scene geometry is encoded via a multi-resolution hash grid, following the Instant-NGP style with 16 levels spanning resolutions $16$--$2048$, a hash table of size $2^{19}$, and 2 features per level. A base MLP, consisting of two layers and 64 hidden units, maps hash-encoded positions to a scalar density $\sigma$ and a 15-dimensional geometry feature vector. A dedicated semantic head, implemented as separate MLP branch with 2 layers and 64 hidden units, takes the geometry features as input and outputs a $C$-dimensional logit vector per sample. Semantic gradients are detached from the density backbone, ensuring that the semantic task does not corrupt geometry learning. Per-pixel semantic predictions are obtained by accumulating per-sample logits along each ray using the same density-derived weights as RGB rendering, producing a rendered $C$-dim logit vector per pixel.

The model is trained for 30\,000 iterations with the combined loss
\begin{equation}
    \mathcal{L} = \mathcal{L}_{\text{rgb}} + \lambda\,\mathcal{L}_{\text{sem}} + \mathcal{L}_{\text{interlevel}} + \mathcal{L}_{\text{distortion}}
\end{equation}
where $\mathcal{L}_{\text{rgb}}$ is the MSE photometric loss, $\mathcal{L}_{\text{sem}}$ is the cross-entropy between rendered logits and semantic labels, and the remaining terms regularize the proposal networks. $\mathcal{L}_{\text{sem}}$ is weighted with class-weights $\lambda{=}[0,\, 1.0,\, 1.2,\, 1.0]$ for [background, leaf, stem, pot] effectively excluding background pixels from the semantic loss, while slightly up-weighting stems to increase focus on the thin structures. We use 4\,096 rays per batch, Adam with learning rate $10^{-2} \to 10^{-4}$ (cosine schedule, 500-step warmup), mixed-precision training (FP16 forward pass, FP32 gradients) and camera pose refinement (SO3${\times}$R3, lr${}=6{\times}10^{-4}$). Training one scene takes approximately 35 minutes on a single NVIDIA A5000 GPU.

\subsection{Point Cloud Extraction and Post-Processing}
\label{sec:pc_export}

We extract the final semantic point cloud via depth backprojection from the training cameras rather than dense volumetric grid sampling. For every second training view at half resolution, the trained NeRF renders per-pixel depth, RGB, accumulation, and semantic logits. Each pixel is back-projected to 3D as $\mathbf{p} = \mathbf{o} + t\,\mathbf{d}$, where $\mathbf{o}$ and $\mathbf{d}$ are the ray origin and direction and $t$ is the rendered depth. A four-stage mask retains only reliable foreground points: (i)~accumulation $> 0.2$ removes uncertain regions, (ii)~$\operatorname{argmax}$ semantic class $\geq 1$ discards background pixels, (iii)~a depth-gradient filter ($|\nabla d| < 0.02$) suppresses bleeding at depth discontinuities, and (iv)~a per-scene axis-aligned bounding box removes outliers. Per-point class labels are obtained as the $\operatorname{argmax}$ of the rendered semantic logits. The resulting cloud contains exactly 5\,M foreground points per scene before post-processing. 

Prior to evaluation, all predicted 3D point clouds undergo identical post-processing to remove NeRF sampling artifacts. First, a connected component filter is applied and only the largest component is kept, to remove floating artifacts. Subsequently, statistical outlier removal eliminates residual sparse noise by discarding points whose mean distance to their $k{=}6$ nearest neighbors exceeds global mean $+ 2\sigma$. 

\section{Experimental Setup}
\label{sec:experiments}

\subsection{Datasets}
\label{sec:datasets}

\noindent\textbf{Begonia dataset.}\quad
We capture a single \textit{Begonia maculata} specimen using a phenotyping cabinet, which provides six synchronized, calibrated RGB cameras at $1600{\times}1200$\,px resolution. A subset of 150 images was manually annotated at pixel level with view-consistent instance IDs across all images. For the experiments in this paper, the instances are grouped into basic semantic classes \emph{leaf}, \emph{stem} and \emph{pot}. All remaining pixels receive the label \emph{background}. Petioles are mapped to \emph{stem}, visible soil and substrate are mapped to \emph{pot}. A 3D ground-truth point cloud (1.5\,M points) was obtained by training default Nerfacto \cite{tancik_nerfstudio_2023}, followed by manual ground-truth segmentation. We also use manually segmented, pixel-perfect 2D masks as an upper-bound reference for our pipeline, which is further referred to as \textit{oracle}.

\noindent\textbf{Diverse dataset.}\quad
We further evaluate on a dataset from an automated 3D phenotyping platform 
~\cite{pietrzyk_seed_2026}, which uses seven calibrated cameras (101 images per plant) and covers 10 species spanning diverse morphologies. Since no 2D pixel-level ground-truth is available for this dataset, evaluation is performed exclusively in 3D. The pipeline-reconstructed point clouds (5\,M points) are manually annotated by a trained annotator. Since ground truth is annotated on the reconstructed cloud, reconstruction/geometry errors (e.g., missing thin stems) are not penalized by this protocol. Our reported IoU therefore reflects semantic labeling quality on reconstructed points only.
\subsection{Evaluation Protocol}
\label{sec:eval_protocol}
SAM3 labels are compared per frame against the pixel-level ground-truth. We report per-class IoU, precision, recall, and F1, as well as foreground mean IoU (mIoU\textsubscript{fg}) and overall pixel accuracy. The predicted point cloud (4.5\,M points after post-processing) is denser than the ground-truth cloud (1.5\,M points). We therefore evaluate by assigning each ground-truth point the semantic label of its nearest neighbor in the predicted cloud. Since ground-truth is annotated directly on the pipeline-produced point cloud, prediction and reference share identical point sets. Labels are compared per-point without spatial matching. Note that 2D and 3D mIoU are computed on different domains (pixels vs. points) and the reported deltas are indicative trends rather than like-for-like comparisons. To separate NeRF reconstruction errors from 2D segmentation errors, we train the semantic NeRF with pixel-perfect ground-truth masks. These \textit{oracle} results represent the architectural upper bound for the current pipeline configuration.

For both datasets, petioles are classified as \emph{stem}. For species with compound leaves (e.g., Tomato), the rachis is labeled as \emph{stem}, with the label boundary placed at the petiole-lamina junction. Plant support stakes present in two scans (Tomato, Sweet Potato) are labeled as stem by both the pipeline and the annotator. This requires filtering for downstream trait analysis. Annotation variability at stem-leaf boundaries may account for 1--3\,pp IoU variation, particularly for the stem class.

\section{Results}

\subsection{2D Semantic Segmentation on Begonia}
\label{sec:exp_begonia_2d}

To investigate the downstream impact of 2D segmentation quality, we evaluate SAM3~\cite{carion_sam_2025} and Grounded-SAM~\cite{ren_grounded_2024} on the multi-view Begonia dataset (150 images). Both foundation models receive identical text prompts (\texttt{"leaf"}, \texttt{"stem"}, \texttt{"pot"}) and unsegmented pixels receive the label \emph{background}. Grounded-SAM first detects bounding boxes from text prompts via Grounding DINO~\cite{liu_grounding_2024}, then refines them into segmentation masks using SAM~\cite{kirillov_segment_2023}. For both steps we use default model weights and confidence thresholds ($\tau = 0.35$ for boxes, $\tau = 0.25$ for text). SAM3 combines detection and segmentation within one shared architecture and was used with default weights and $\tau = 0.5$.

\begin{table}[t]
\centering
\setlength{\tabcolsep}{4pt}
\begin{tabular}{l l c c c c}
\toprule
Method & Class & IoU & Precision & Recall & F1 \\
\midrule
\multirow{3}{*}{Grounded-SAM~\cite{ren_grounded_2024}}
 & Leaf        & 0.785 & 0.981 & 0.797 & 0.880 \\
 & Stem        & 0.315 & 0.420 & 0.558 & 0.479 \\
 & Pot         & 0.699 & 0.815 & 0.831 & 0.823 \\
\cmidrule{2-6}
 & \textit{mIoU\textsubscript{fg}\,/\,Pixel Acc.} & \multicolumn{4}{c}{\textit{0.600\,/\,0.949}} \\
\midrule
\multirow{3}{*}{\textbf{SAM3}~\cite{carion_sam_2025}}
 & Leaf        & \textbf{0.952} & 0.986 & 0.964 & 0.975 \\
 & Stem        & \textbf{0.705} & 0.972 & 0.720 & 0.827 \\
 & Pot         & \textbf{0.853} & 0.974 & 0.873 & 0.921 \\
\cmidrule{2-6}
 & \textit{mIoU\textsubscript{fg}\,/\,Pixel Acc.} & \multicolumn{4}{c}{\textit{\textbf{0.837}\,/\,\textbf{0.982}}} \\
\bottomrule
\end{tabular}
\caption{Zero-shot 2D semantic segmentation on the Begonia dataset (150 images). Both methods use identical text prompts and default configurations. SAM3 outperforms Grounded-SAM by +23.7\,pp mIoU\textsubscript{fg}, with the largest gap on stems (+39.0\,pp).}
\label{tab:2d_comparison}
\end{table}

Table~\ref{tab:2d_comparison} shows that SAM3 substantially outperforms Grounded-SAM across all classes, achieving +23.7\,pp higher foreground mIoU (0.84 vs.\ 0.60). The improvement is most pronounced for stems (+39\,pp), where Grounded-SAM's two-stage pipeline suffers from compounding errors between detection and segmentation. For leaves, SAM3 achieves near-perfect segmentation (IoU = 0.95) while Grounded-SAM reaches only 0.79, frequently under-segmenting partially occluded leaves that fall outside detected bounding boxes.

\begin{figure}[tbp]
    \centering
    \includegraphics[width=\linewidth]{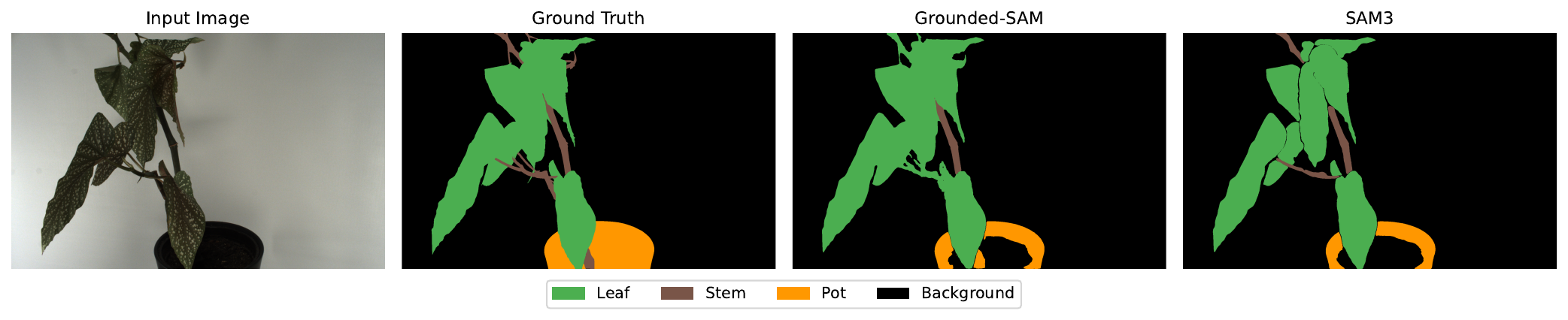}
    \caption{2D segmentation comparison on a representative Begonia image. Grounded-SAM exhibits cross-class confusion, misclassifying stem regions as leaf, whereas SAM3 produces clean masks with the error being missed regions only.}
    \label{fig:mask_comparison}
\end{figure}

The precision and recall columns reveal a key difference in error characteristics. SAM3 exhibits uniformly high precision ($>$0.97) across all classes, with recall being the sole limiting factor (72\% for stems). Errors are overwhelmingly due to missed detections rather than wrong labels. This is the ideal error mode for our NeRF pipeline, since missed regions in one view are compensated by correct detections in other views. Grounded-SAM, by contrast, suffers from cross-class confusion: stem precision of 42\% means that more than half of pixels predicted as stem actually belong to other classes (Fig.~\ref{fig:mask_comparison}). These systematic misclassifications are more damaging for multi-view fusion, as they provide consistently wrong supervisory signal across views rather than recoverable gaps. These results demonstrate how recent advances in unified vision-language architectures strengthen their viability as annotation-free backbones for 3D applications.

\subsection{3D Semantic Segmentation on Begonia}
\label{sec:exp_begonia_3d}

\begin{figure}[t!]
  \centering
  \setlength{\tabcolsep}{5pt}
  \begin{tabular}{l c c c c c}
    \toprule
    2D Mask Source & mIoU & Acc. & IoU\textsubscript{leaf} 
                   & IoU\textsubscript{stem} & IoU\textsubscript{pot} \\
    \midrule
    Oracle (GT masks)              & 0.966 & 0.991 & 0.987 & 0.915 & 0.995 \\
    \midrule
    Grounded-SAM~\cite{ren_grounded_2024} 
                                   & 0.792 & 0.923 & 0.890 & 0.508 & 0.979 \\
    \textbf{SAM3}~\cite{carion_sam_2025}  
                                   & \textbf{0.926} & \textbf{0.980} 
                                   & \textbf{0.984} & \textbf{0.829} & 0.964 \\
    \midrule
    \rowcolor[gray]{0.93}
    $\Delta$\,(G-SAM 3D $-$ G-SAM 2D) & \textit{+19.2} & --- 
                                    & \textit{+10.5} & \textit{+19.3} 
                                    & \textit{+28.0} \\
    \rowcolor[gray]{0.93}
    $\Delta$\,(SAM3 3D $-$ SAM3 2D) & \textit{+8.9} & --- 
                                    & \textit{+3.2} & \textit{+12.4} 
                                    & \textit{+11.1} \\
    \rowcolor[gray]{0.93}
    $\Delta$\,(Oracle $-$ SAM3 3D)  & \textit{+4.0} & --- 
                                    & \textit{+0.3} & \textit{+8.6}  
                                    & \textit{+3.1} \\
    \bottomrule
  \end{tabular}
  \captionof{table}{3D semantic segmentation on the Begonia dataset. The two
$\Delta$ rows in the middle quantify the multi-view consensus gain over per-frame
2D predictions (Table~\ref{tab:2d_comparison}) for both mask sources. The last row
shows the remaining gap to perfect 2D input (oracle). SAM3 outperforms
Grounded-SAM by +13.4\,pp mIoU and reaches 95.9\% of the oracle performance
ceiling.}
  \label{tab:3d_competition}

  \vspace{6pt}  
  \includegraphics[width=0.3\linewidth]{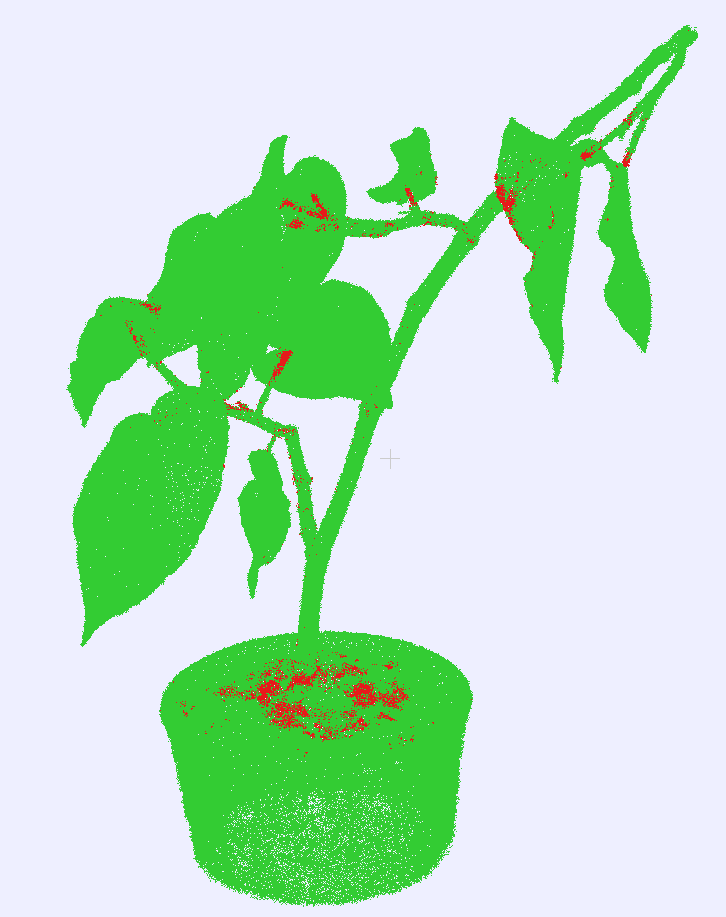}
  \captionof{figure}{Colored segmentation errors of the SAM3 masked-based run for the Begonia point cloud. Correctly segmented points are green, points with the wrong label red.}
  \label{fig:zeroshot_begonia_errors}
\end{figure}

Following the investigation of the 2D masks, this section presents the main contribution of this paper: \textit{Is lifting of the labels of 2D foundation models into 3D a valid approach for 3D plant organ segmentation?}
To answer this, we evaluate the full pipeline by comparing semantic point clouds against manually annotated 3D ground truth (1.5\,M points) of the Begonia dataset. Table~\ref{tab:3d_competition} reports results for three 2D mask sources fed into our pipeline.

Lifting SAM3 masks into the semantic NeRF improves mIoU by +8.9\,pp over per-frame 2D predictions. The gain is largest for stems (+12.4\,pp), where single-view predictions suffer from occlusion and thin geometry. Leaves, already near-perfect in 2D (0.952), still improve to 0.984 in 3D, demonstrating that even small per-frame errors are corrected through multi-view redundancy. Consistent observations across multiple viewpoints allow the NeRF to recover correct labels through implicit majority voting. \textbf{This confirms our hypothesis that imperfect zero-shot 2D masks become high-quality 3D annotations when fused across multiple views.}

Fig.~\ref{fig:zeroshot_begonia_errors} reveals two main error sources in the SAM3 3D segmentation. First, boundaries between stem and leaf are error-prone, an inherent limitation of semantic NeRFs, whose continuous neural field produces smooth rather than sharp class transitions. Second, \textit{pot} region points are frequently misclassified as \textit{stem}, because SAM3 does not segment the substrate as pot (see Fig. \ref{fig:mask_comparison}), leaving the NeRF with ambiguous information that propagates into mixed stem/pot predictions.

Grounded-SAM also benefits substantially from multi-view fusion (+19.2\,pp mIoU), demonstrating that the consensus mechanism is model-agnostic. The larger absolute gain compared to SAM3 (+8.9\,pp) can be attributed to more room for improvement from noisier 2D masks. However, the final 3D result remains 13.4\,pp below SAM3. Stem IoU improves from 0.315 to 0.508. The systematic cross-class confusion identified in Sec.~\ref{sec:exp_begonia_2d} is only partially resolved through multi-view fusion. When misclassifications are consistent across views rather than random, they provide coherent wrong signal that the NeRF cannot override by majority vote. Pot IoU (0.979) benefits from the fact that stem-labeled pixels at pot-stem boundaries are fewer than pot-labeled pixels, so the NeRF resolves boundary ambiguity in its favor.

The oracle achieves 96.6\% mIoU, establishing the architectural upper bound for our NeRF configuration. Relative to this upper bound only 4.0\,pp of the total 7.4\,pp gap between zero-shot performance and perfection are attributable to SAM3 mask imperfections (mostly stems with a 8.6\,pp gap vs.\ 0.3\,pp for leaves). The remaining 3.4\,pp reflect inherent NeRF limitations: the continuous neural field smooths class boundaries at organ junctions rather than producing sharp semantic transitions. This suggests that future improvements should prioritize boundary-aware NeRF architectures over 2D segmentation refinement.

\subsubsection{Confidence Threshold Ablation}
\label{sec:conf_sweep}

\begin{figure}
    \centering
    \includegraphics[width=1\linewidth]{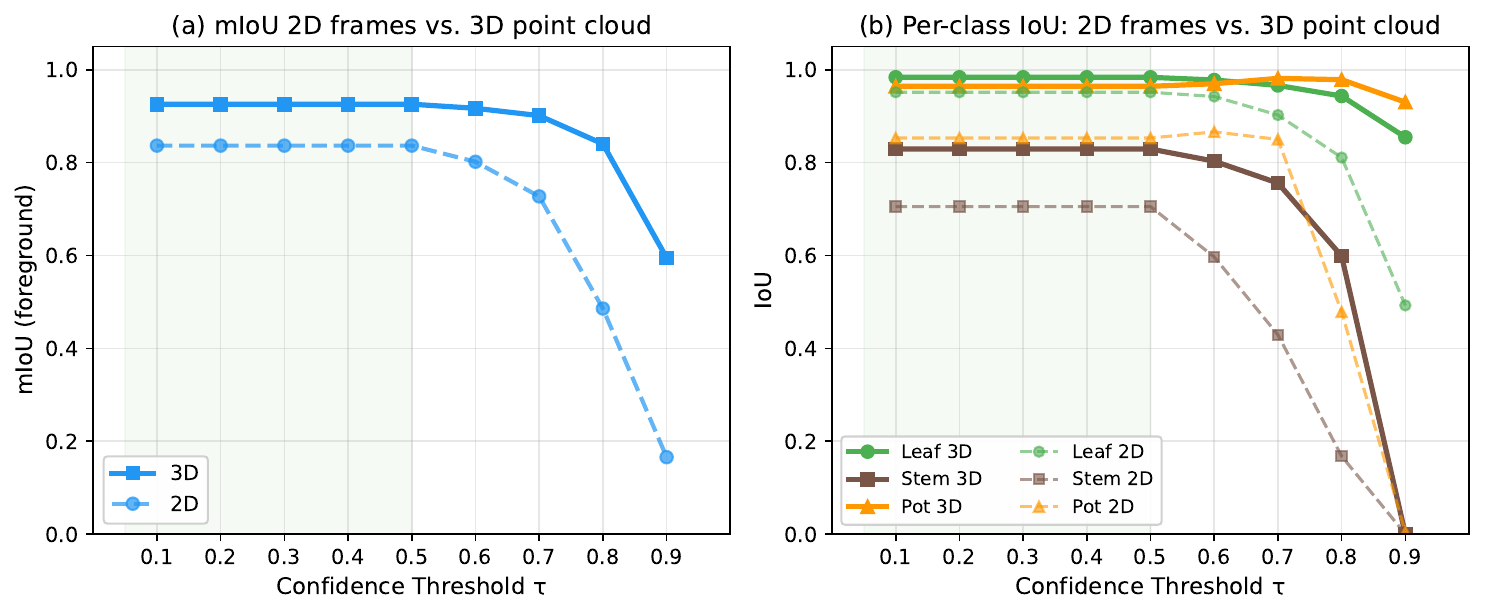}
    \caption{
    Confidence threshold analysis. Solid lines show 3D point cloud metrics of pipeline output. Dotted lines show 2D per-frame predictions. (a)~mIoU is stable for $\tau \leq 0.5$. The 3D curve degrades less due to multi-view redundancy. (b)~Per-class breakdown: 3D consistently improves over 2D, with stems as the primary bottleneck collapsing at high thresholds.}
    \label{fig:conf_sweep}
\end{figure}

In our preliminary, unreported experiments with Grounded-SAM, we observed that increasing the confidence threshold improved downstream 3D results by suppressing conflicting cross-class predictions. These findings motivated our ablation studies with SAM3. We sweep $\tau \in \{0.1, \ldots, 0.9\}$ and report 2D mask quality, pipeline-aware metrics (Sec.~\ref{sec:eval_protocol}), and downstream 3D segmentation in Fig.~\ref{fig:conf_sweep}.

Results are \emph{identical} for all $\tau \leq 0.5$, revealing bimodal confidence distributions: SAM3 predictions are either highly confident ($>$0.5) or absent. For SAM3, cross-class misclassification stays below 0.8\% at all thresholds. Increasing $\tau$ beyond 0.5 does not improve prediction quality, it only reduces foreground coverage. This inverts the Grounded-SAM behavior that benefited from aggressive filtering to suppress its frequent cross-class confusions.

\subsection{Cross-Species Generalization}
\label{sec:sepp}

To evaluate whether our pipeline generalizes beyond the single Begonia specimen, we apply it to ten species from an automated multi-view phenotyping platform
\cite{pietrzyk_seed_2026}. All species are processed with unchanged pipeline settings and text prompts (\texttt{"leaf"}, \texttt{"stem"}, \texttt{"pot"}). To investigate zero-shot extensibility, two species receive one additional prompt each (\texttt{"dead leaf"} for Sweet Potato, \texttt{"flower"} for Strawberry). 

\begin{table}[t]
\centering
\small
\setlength{\tabcolsep}{3pt}
\begin{tabular}{l c c c c c c}
\toprule
Species & mIoU & Acc. & IoU\textsubscript{leaf} & IoU\textsubscript{stem} & IoU\textsubscript{pot} & IoU \\
\midrule
Norway Maple     & 0.970 & 0.994 & 0.993 & 0.918 & 0.997 & --- \\
Sycamore Maple   & 0.953 & 0.993 & 0.993 & 0.873 & 0.993 & --- \\
Linden           & 0.946 & 0.986 & 0.987 & 0.878 & 0.973 & --- \\
Sunflower        & 0.922 & 0.985 & 0.978 & 0.795 & 0.994 & --- \\
Sweet Potato     & 0.899 & 0.982 & 0.974 & 0.909 & 0.985 & 0.729\textsuperscript{$\dagger$} \\
Geranium         & 0.879 & 0.966 & 0.951 & 0.728 & 0.959 & --- \\
Tomato           & 0.833 & 0.942 & 0.947 & 0.645 & 0.907 & --- \\
Strawberry       & 0.727 & 0.933 & 0.963 & 0.358 & 0.919 & 0.669\textsuperscript{$\ddagger$} \\
Lemon            & 0.725 & 0.929 & 0.919 & 0.330 & 0.923 & --- \\
Pepper           & 0.707 & 0.945 & 0.962 & 0.231 & 0.928 & --- \\
\midrule
\textit{Mean $\pm$ std} & \textit{0.856$\pm$0.097} & \textit{0.965} & \textit{0.967$\pm$0.022} & \textit{0.667$\pm$0.251} & \textit{0.958$\pm$0.033} & \\
\bottomrule
\multicolumn{7}{l}{\footnotesize \textsuperscript{$\dagger$}dead leaf \quad \textsuperscript{$\ddagger$}flower}
\end{tabular}
\caption{Cross-species 3D semantic segmentation on the diverse dataset ($\tau{=}0.5$, no per-species tuning). Our pipeline achieves $\ge$0.90 mIoU on four species and $\mathbf{0.856}$ mean mIoU.}
\label{tab:sepp_results}
\end{table}

Table~\ref{tab:sepp_results} shows that the pipeline generalizes across all ten species with a mean mIoU of $0.856\pm0.097$. Leaf IoU exceeds 0.91 (mean 0.967) and pot IoU exceeds 0.90 (mean 0.958) for \emph{every} species tested, regardless of morphological differences. This demonstrates that SAM3's text-conditioned segmentation, when lifted through multi-view consensus, provides robust organ separation across diverse plant architectures without any species-specific adaptation.

Stem IoU exhibits the highest variance (0.667$\pm$0.251) and is the dominant factor in per-species mIoU differences. A clear pattern emerges: species with simple, clearly separated stem architecture outperform those with dense, tangled branching. This mirrors the Begonia finding (Sec.~\ref{sec:exp_begonia_3d}) where stems constituted the primary bottleneck. The difficulty is compounded for species where thin stems run parallel to or are partially occluded by leaves and pot edges, creating ambiguous regions in both 2D masks and 3D reconstruction.

Adding a single text prompt per novel class is sufficient to segment additional organ types. Sweet Potato's dead leaves (IoU = 0.729) and Strawberry's flowers (IoU = 0.669) are detected without modifying any pipeline component. While these achieve lower IoU than the core classes, they demonstrate that the pipeline's vocabulary is trivially extensible at inference time.

The lowest-performing species (Pepper, Lemon and Strawberry) share a common failure mode: pot points misclassified as stem at the soil-stem emergence zone (Pepper: 198K/3.1M pot points and Strawberry: 182K/2.4M). Limited camera coverage of the plant emergence point region combined with the slightly elevated stem class weight ($\lambda_{\text{stem}} = 1.2$ vs.\ $\lambda_{\text{pot}} = 1.0$) and no masks for the pot filling substrate cause the NeRF to resolve the ambiguity in favor of stem. While this decreases stem precision, the resulting false positives form a thin planar layer on the pot surface that is trivial to remove in post-processing.

\begin{figure*}[p]
  \centering
  \setlength{\tabcolsep}{1pt}
  \renewcommand{\arraystretch}{0.9}

  \newcommand{\leafc}{\textcolor[RGB]{0,158,115}{\rule{0.8em}{0.8em}}}
  \newcommand{\stemc}{\textcolor[RGB]{230,159,0}{\rule{0.8em}{0.8em}}}
  \newcommand{\potc}{\textcolor[RGB]{86,180,233}{\rule{0.8em}{0.8em}}}
  \newcommand{\flowerc}{\textcolor[RGB]{204,121,167}{\rule{0.8em}{0.8em}}}
  \newcommand{\deadLeaf}{\textcolor[RGB]{240,227,66}{\rule{0.8em}{0.8em}}}

  \begin{minipage}{\linewidth}
    \centering
    \small\textit{Begonia maculata} (mIoU 0.926)\\[1pt]
    \begin{tabular}{@{}cccc@{}}
      \includegraphics[width=0.215\linewidth]{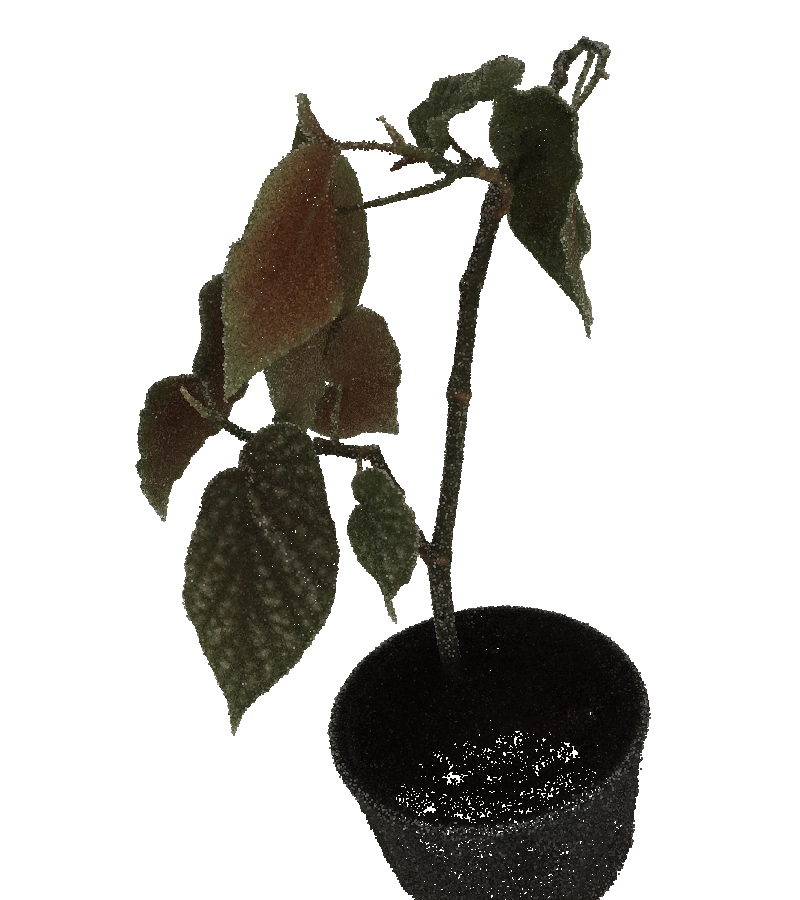} &
      \includegraphics[width=0.215\linewidth]{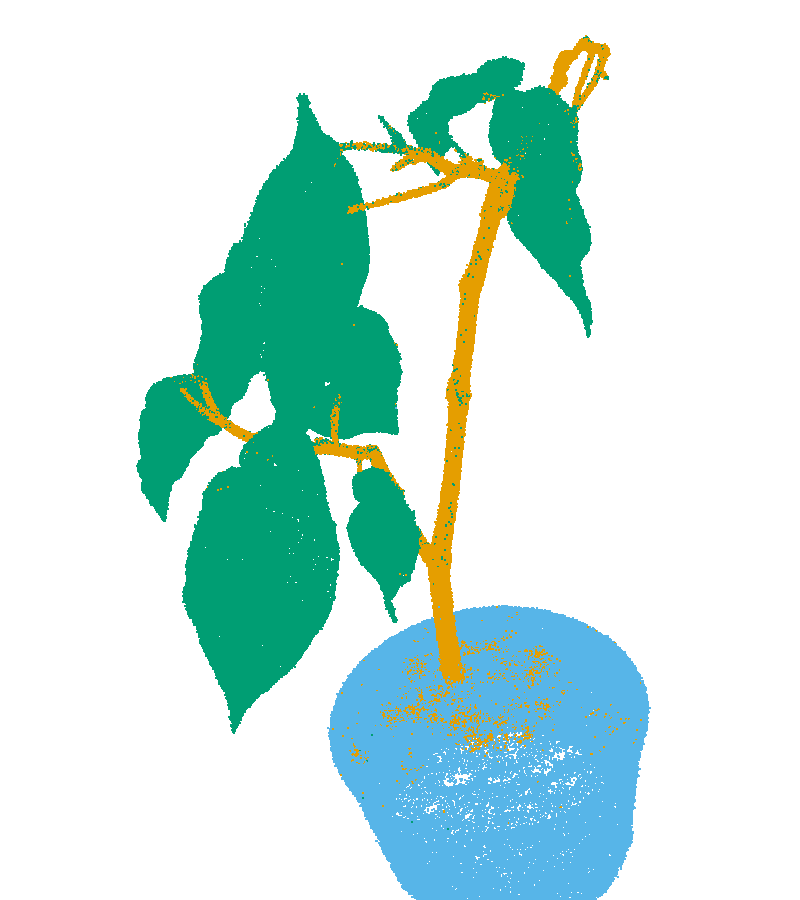} &
      \includegraphics[width=0.215\linewidth]{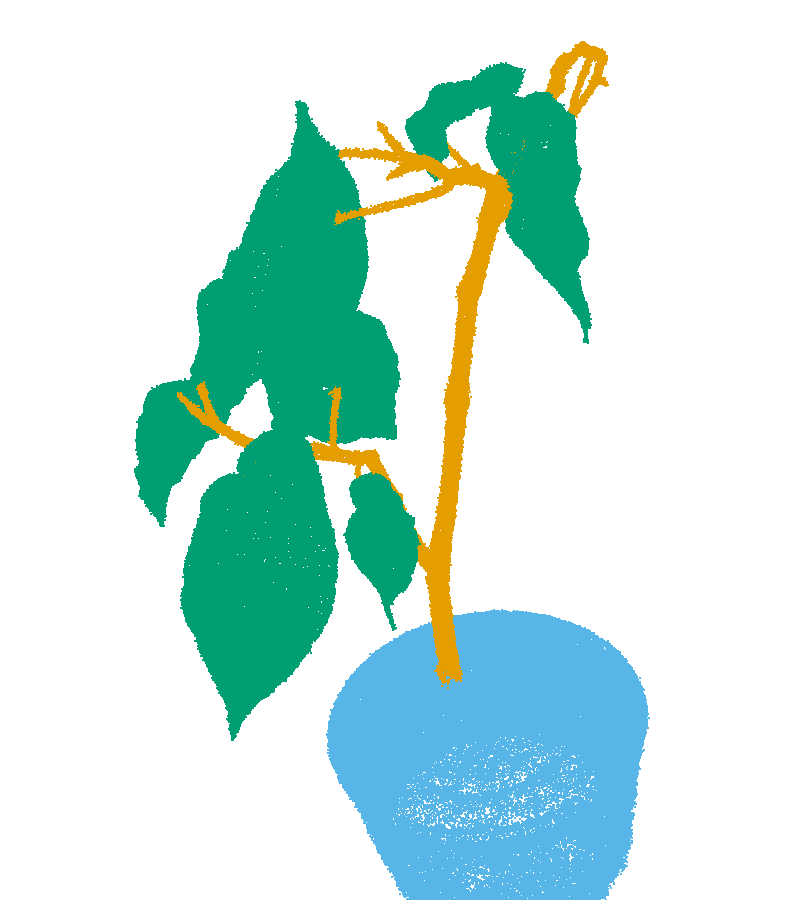} &
      \includegraphics[width=0.215\linewidth]{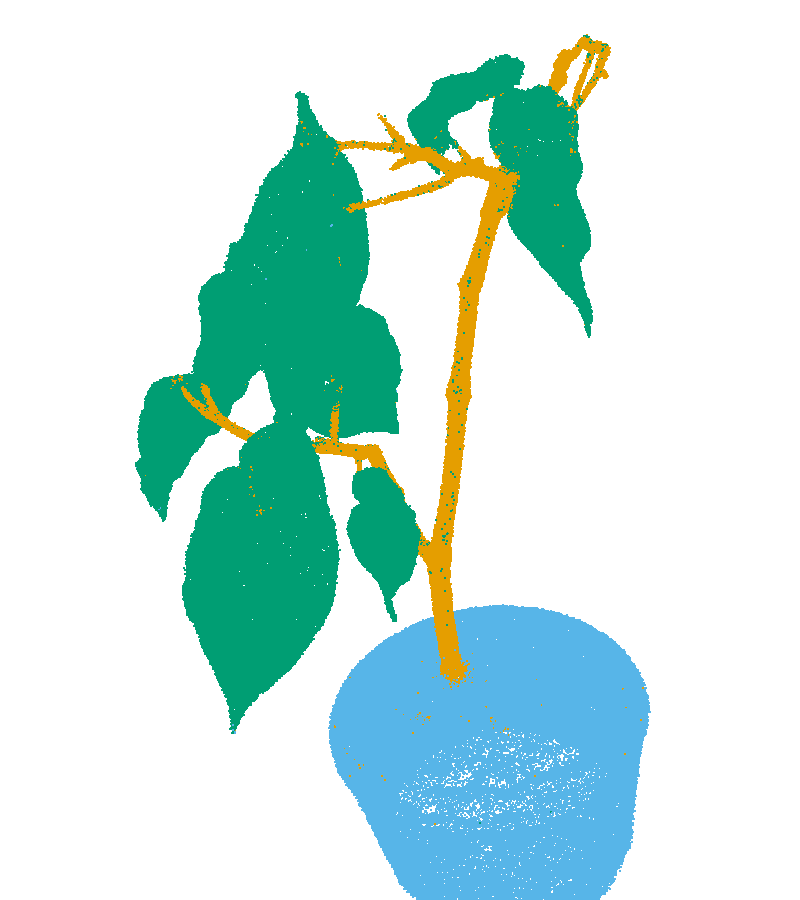} \\
      \footnotesize RGB & \footnotesize Predicted (SAM3) & \footnotesize Ground Truth & \footnotesize Oracle
    \end{tabular}
  \end{minipage}

  \vspace{2pt}
  \hrule height 0.4pt
  \vspace{2pt}

  \begin{tabular}{@{}cccc@{}}
    \multicolumn{2}{c}{\small Norway Maple (mIoU 0.970)} &
    \multicolumn{2}{c}{\small Sunflower (mIoU 0.922)} \\
    \includegraphics[width=0.215\linewidth]{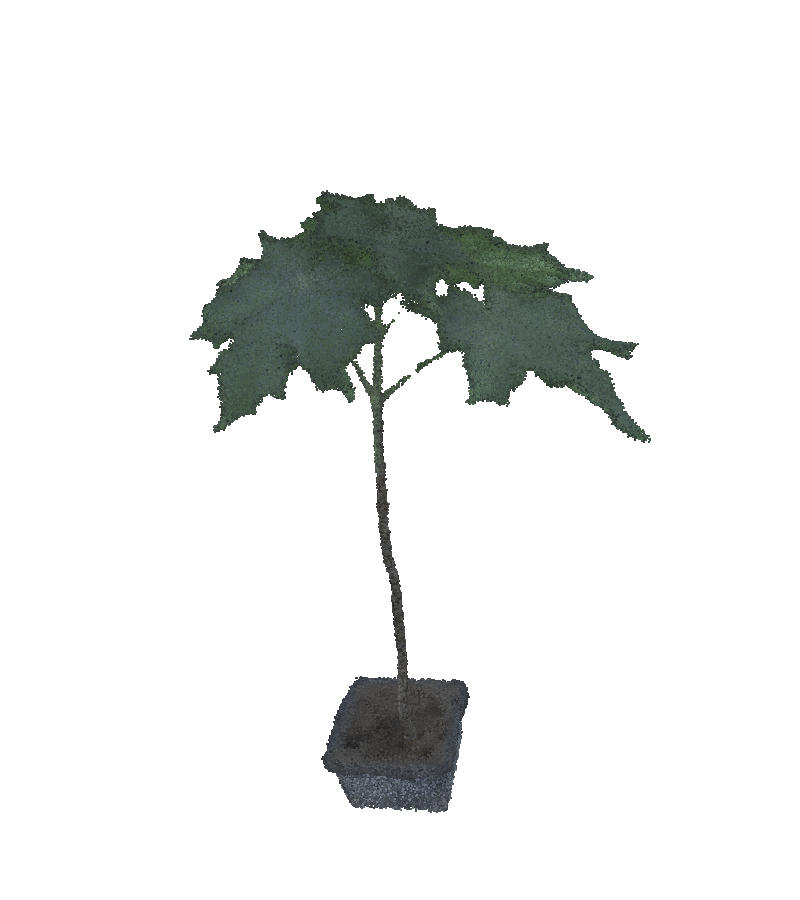} &
    \includegraphics[width=0.215\linewidth]{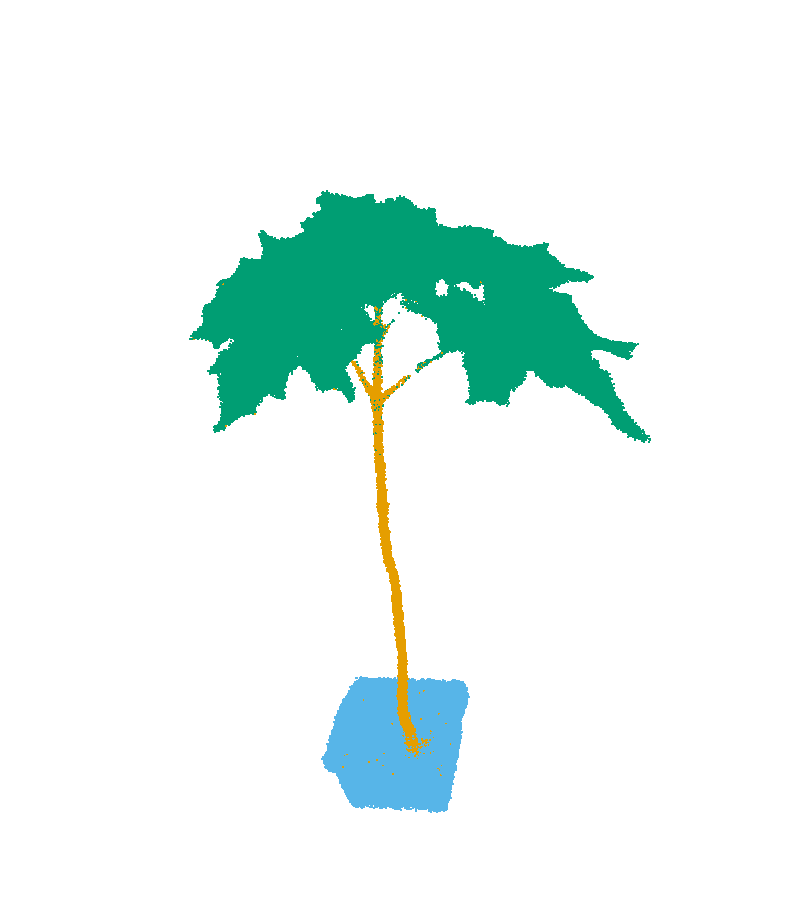} &
    \includegraphics[width=0.215\linewidth]{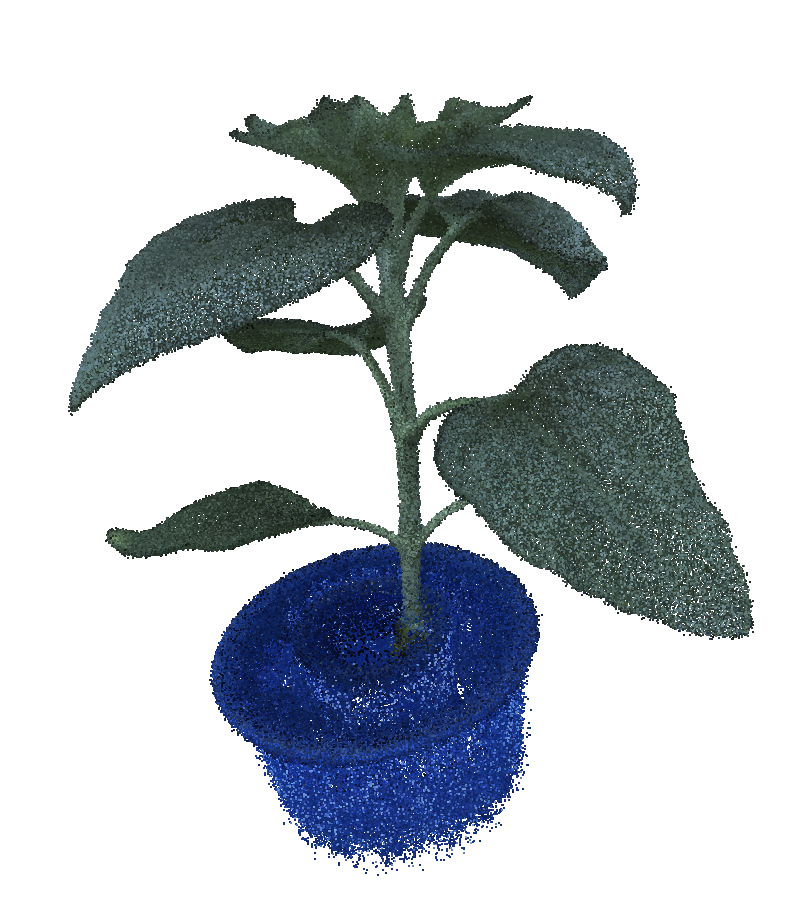} &
    \includegraphics[width=0.215\linewidth]{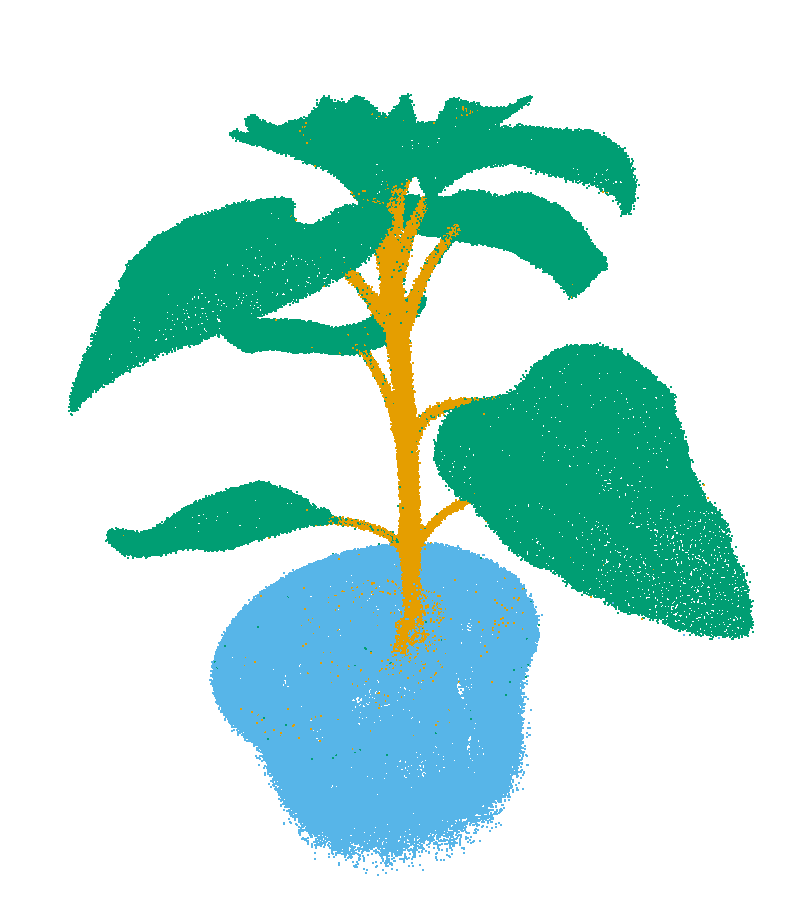} \\

    \multicolumn{2}{c}{\small Geranium (mIoU 0.879)} &
    \multicolumn{2}{c}{\small Tomato (mIoU 0.833)} \\
    \includegraphics[width=0.215\linewidth]{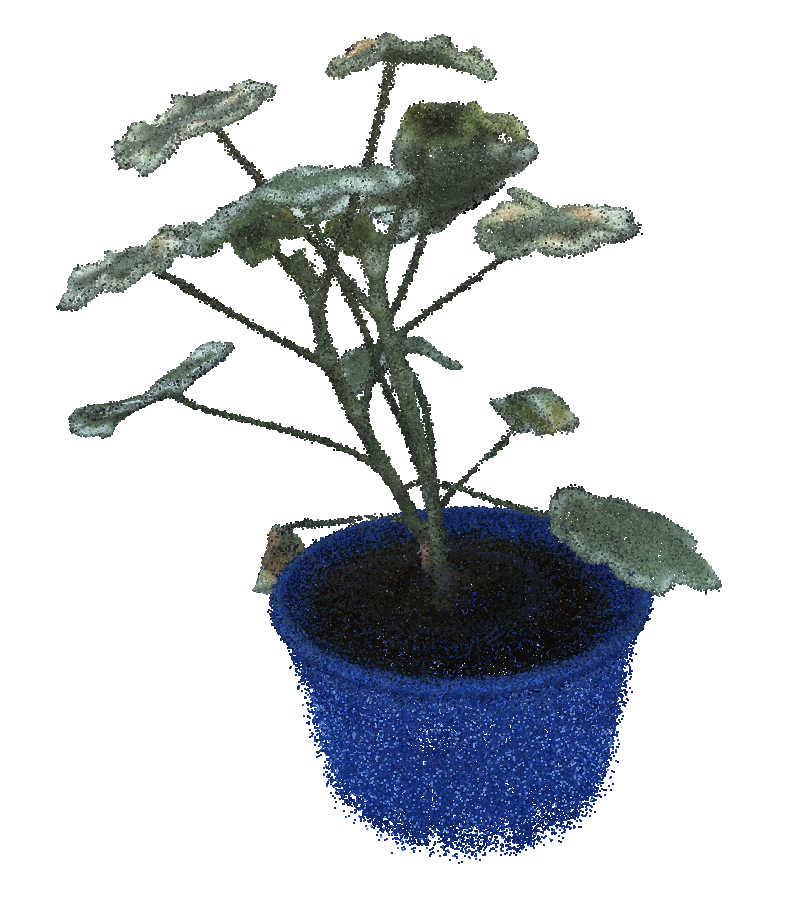} &
    \includegraphics[width=0.215\linewidth]{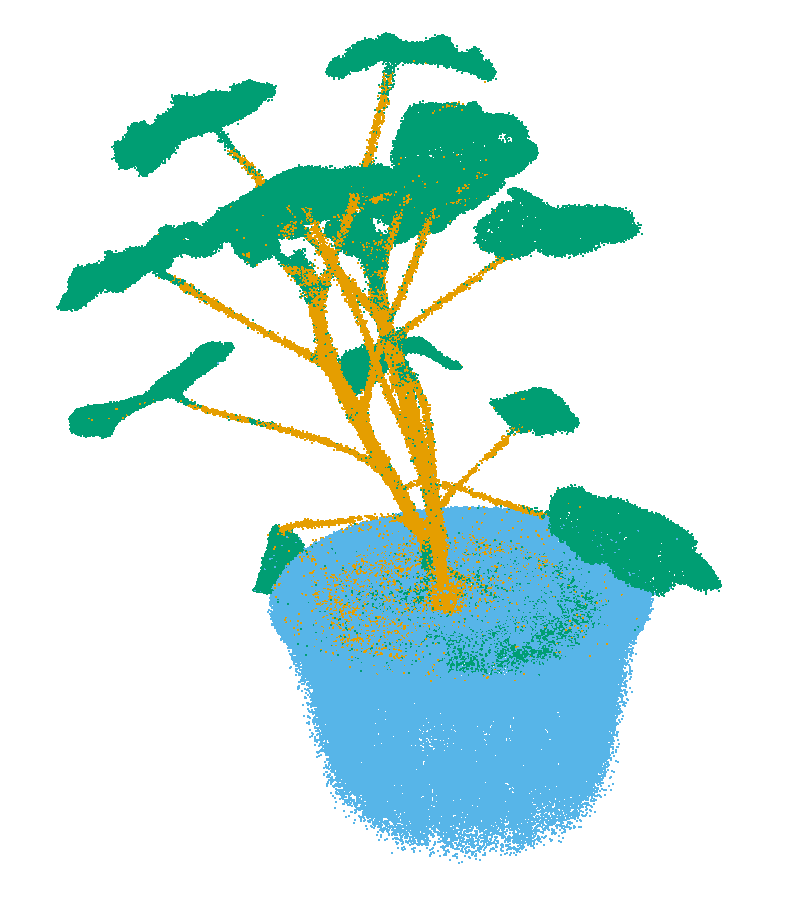} &
    \includegraphics[width=0.215\linewidth]{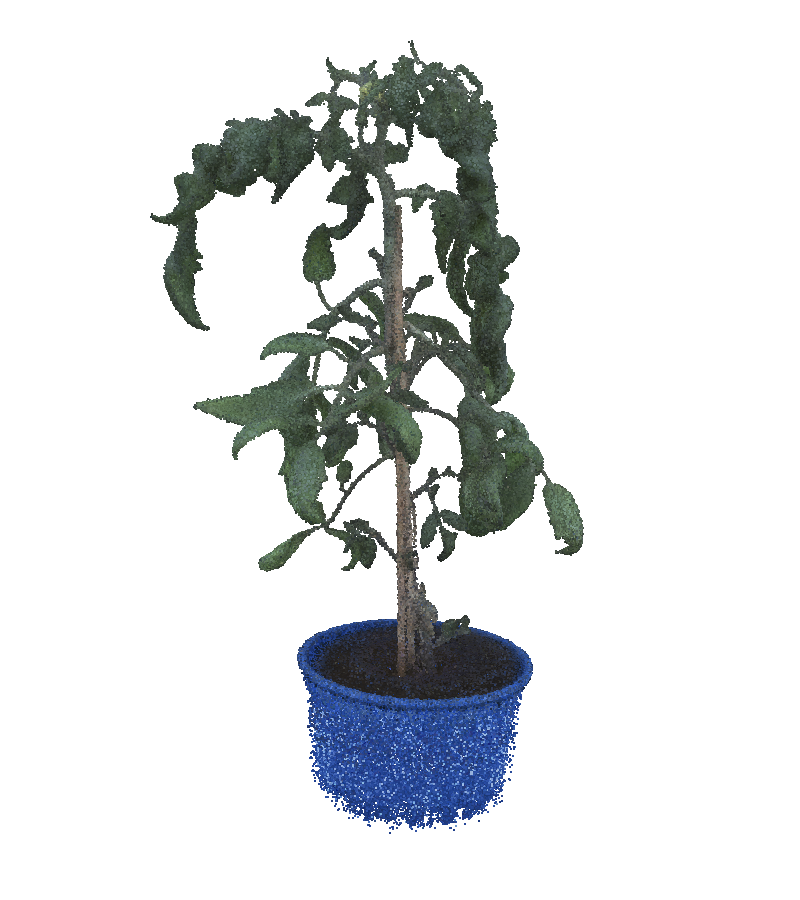} &
    \includegraphics[width=0.215\linewidth]{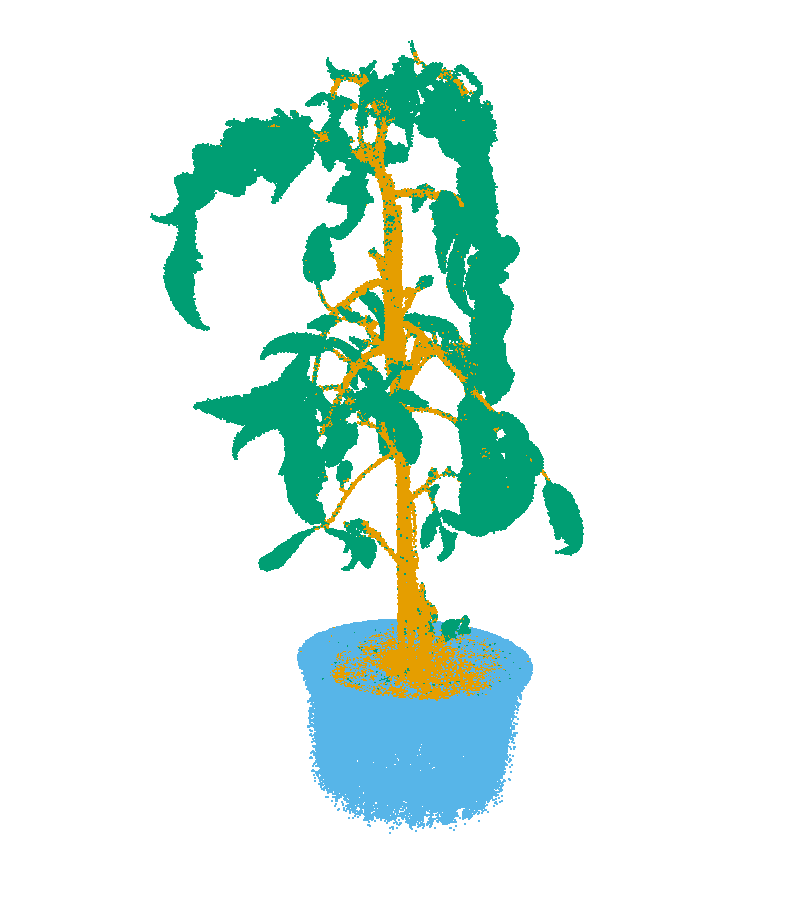} \\

    \multicolumn{2}{c}{\small Pepper (mIoU 0.707)} &
    \multicolumn{2}{c}{\small Lemon (mIoU 0.725)} \\
    \includegraphics[width=0.215\linewidth]{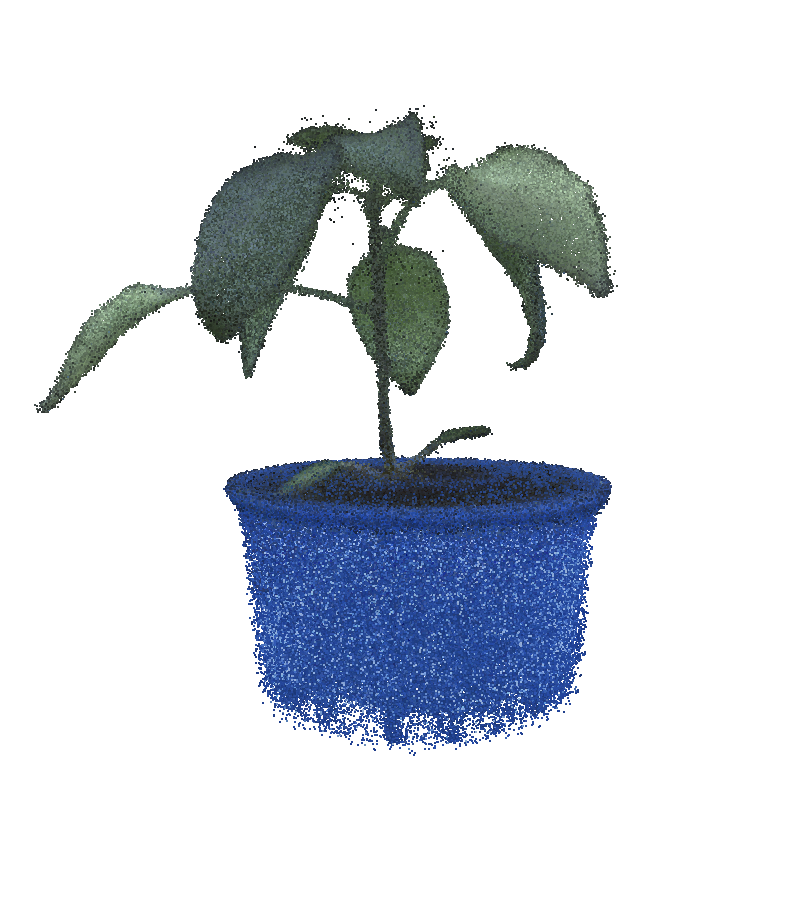} &
    \includegraphics[width=0.215\linewidth]{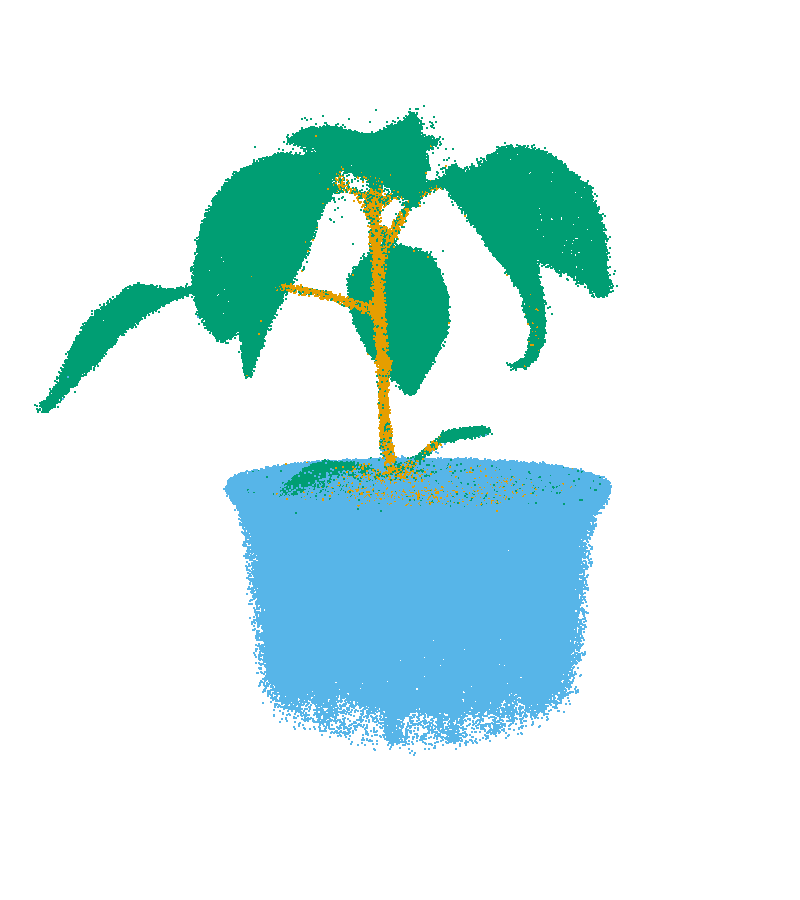} &
    \includegraphics[width=0.215\linewidth]{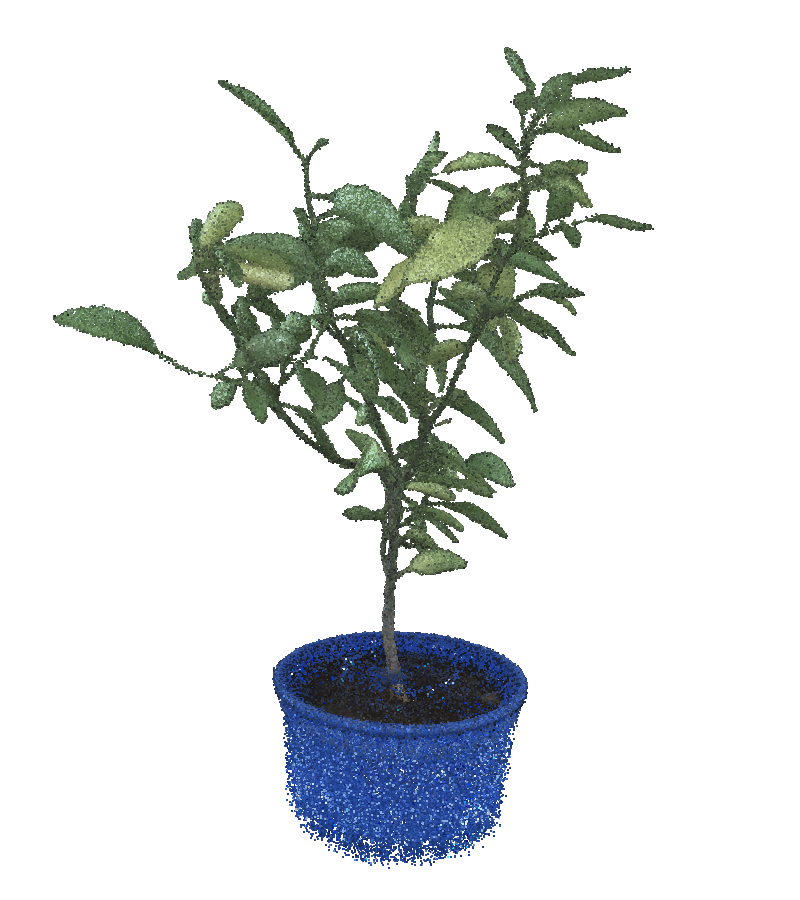} &
    \includegraphics[width=0.215\linewidth]{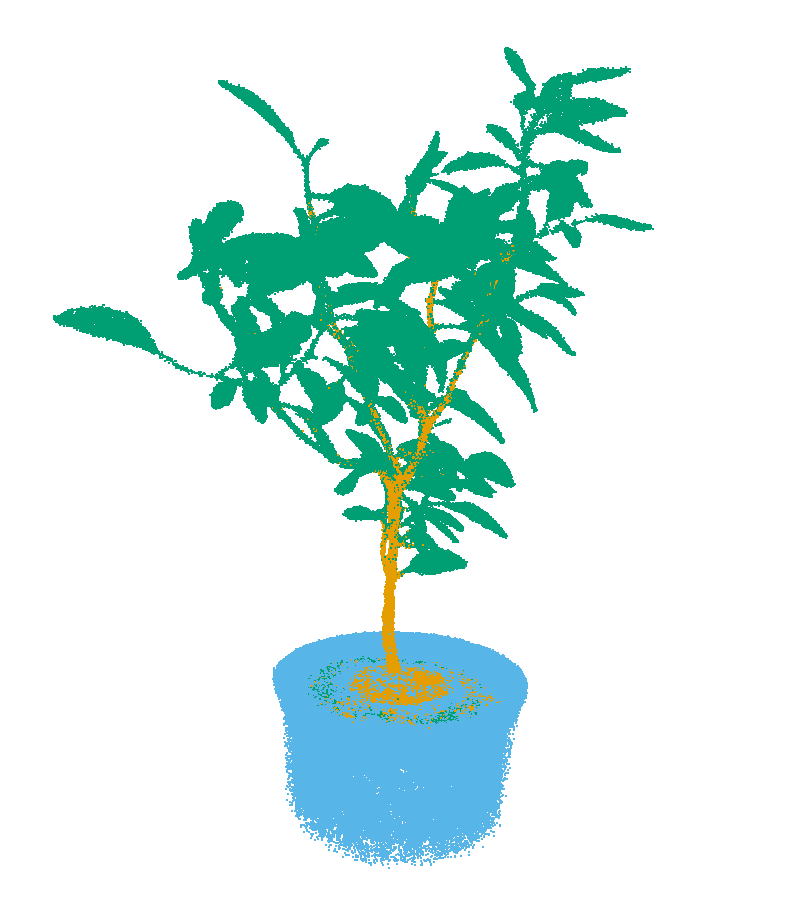} \\

    \multicolumn{2}{c}{\small Sweet Potato (mIoU 0.899)$\dagger$} &
    \multicolumn{2}{c}{\small Strawberry (mIoU 0.727)$\ddagger$} \\
    \includegraphics[width=0.215\linewidth]{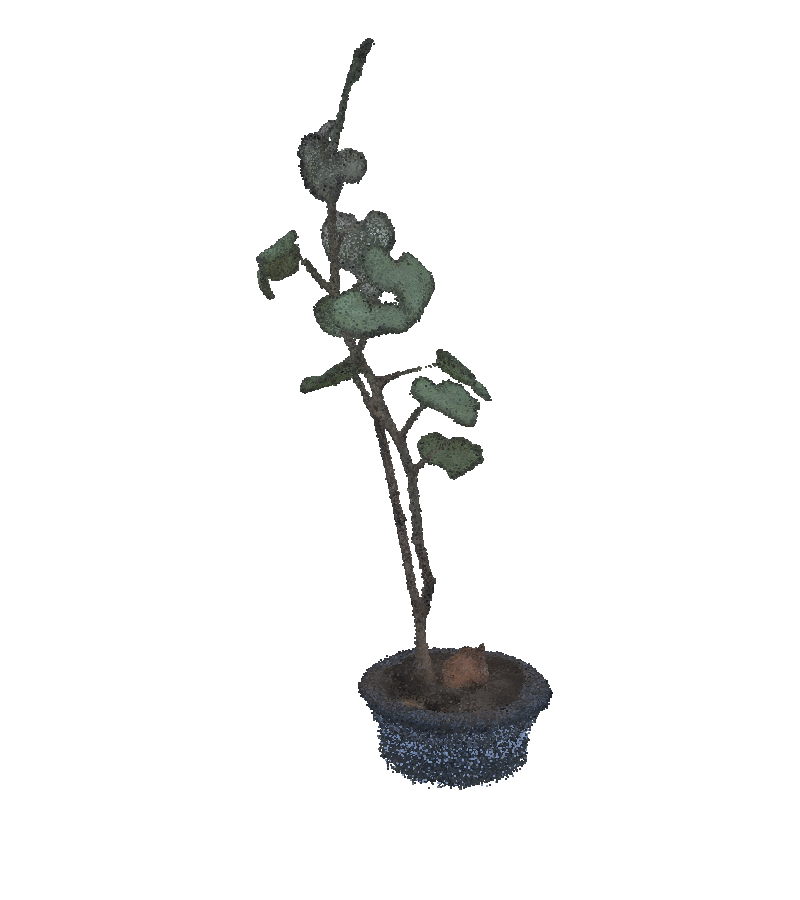} &
    \includegraphics[width=0.215\linewidth]{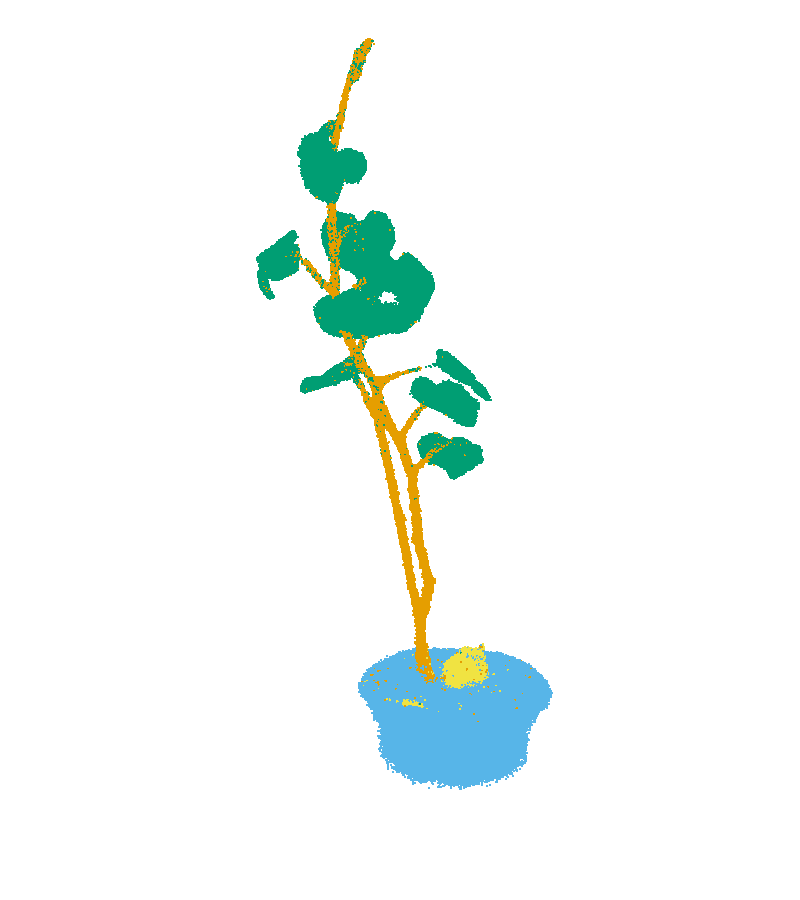} &
    \includegraphics[width=0.215\linewidth]{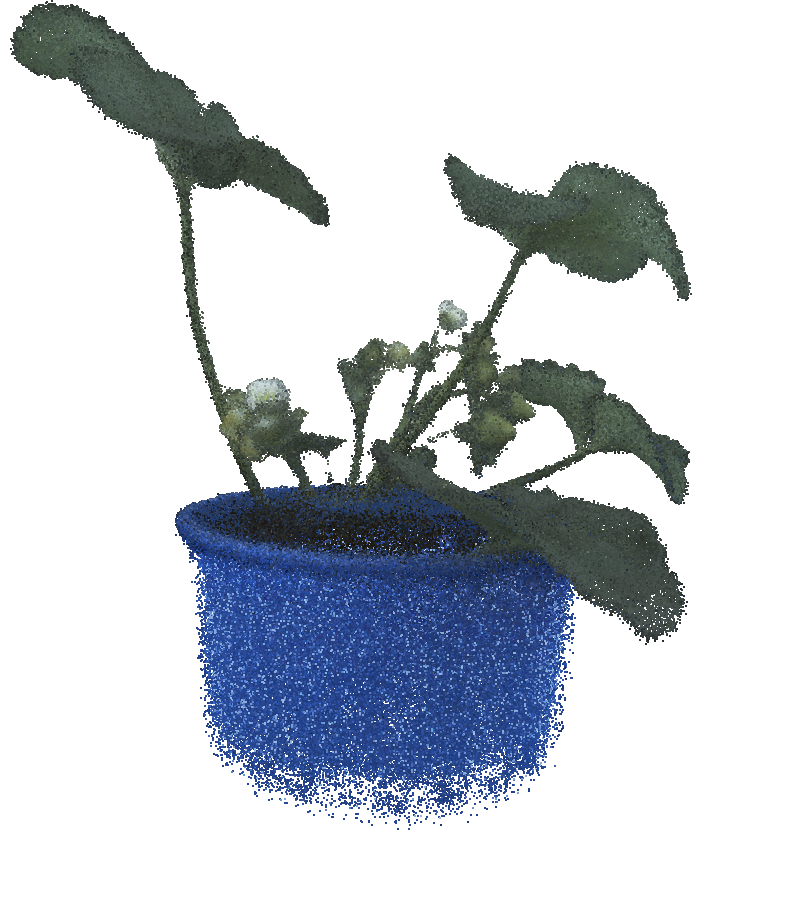} &
    \includegraphics[width=0.215\linewidth]{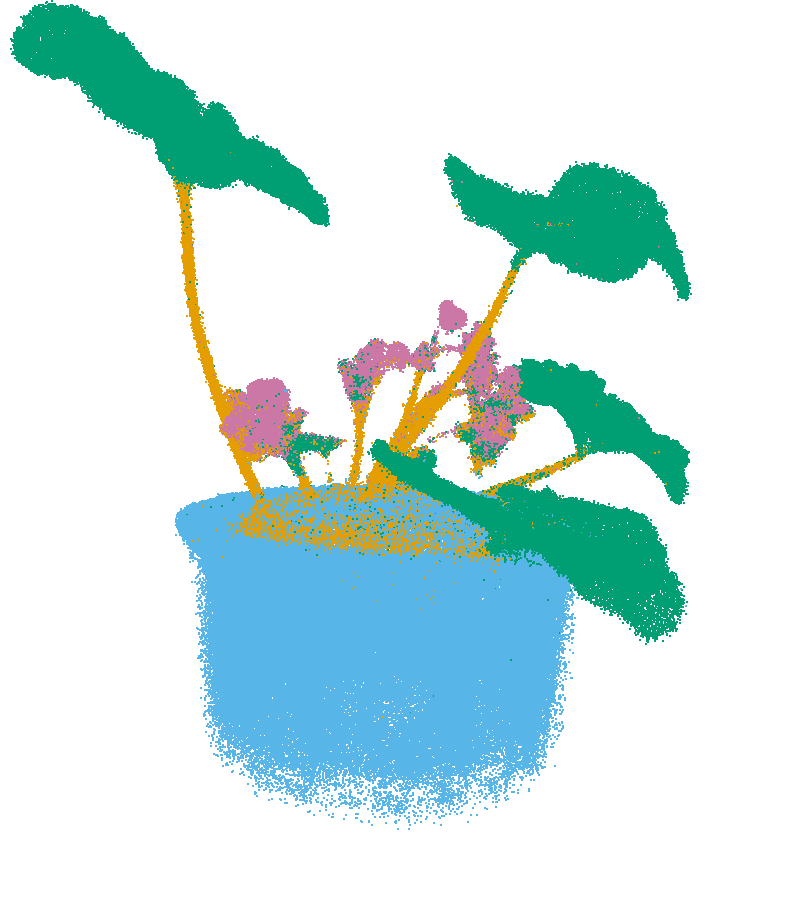} \\
  \end{tabular}

  \vspace{2pt}
  \leafc~Leaf\quad\stemc~Stem\quad\potc~Pot\quad$\dagger$~\deadLeaf~Dead Leaf\quad$\ddagger$~\flowerc~Flower

  \caption{\small Qualitative 3D segmentation results. \textbf{Top:} Begonia with ground-truth comparison. \textbf{Bottom:} Cross-species results with RGB point cloud (left) and predicted semantic point cloud (right) per pair. All predictions use identical pipeline settings without species-specific tuning.}
  \label{fig:hero_3d}
\end{figure*}

\section{Discussion}

The +8.9 pp (SAM3) and +19.2 pp (Grounded-SAM) gains from 2D→3D confirm our hypothesis that multi-view fusion is an effective self-correcting mechanism for lifting 2D labels into 3D. The foundation model used as backbone for our pipeline is a critical element and its error-mode needs to be well understood. SAM3's errors are predominantly missed detections that are recovered via multi-view redundancy. Grounded-SAM's cross-class confusions, by contrast, cause the NeRF to produce wrong classifications. Thus, for our pipeline, precision is more important than recall in 2D masks. The oracle experiments (96.6\% mIoU) show that there remains a small error gap, even when 2D labels are perfect (3.4 pp). However, for the Begonia dataset both SAM3 labels and 2D ground truth are sufficient for getting 3D segmentation results that are usable for phenotyping applications. While the gap remains largest for the stem class (8.6 pp), the SAM3 zero-shot pipeline is very close to the upper boundary for leaf and pot. For further improvements, a prioritization of boundary-aware NeRF representations seems more promising than focusing on better 2D masks. Worse performance on thin structures with visual ambiguity at organ boundaries (e.g, stem--leaf) can also be observed in supervised approaches (e.g., PSegNet, Organ3DNet), especially for dense plants with many occlusions. Stem segmentation remains an open challenge for the 3D phenotyping domain.

In terms of performance our pipelines generalizes well across all eleven (incl. Begonia) morphologically diverse species using identical text prompts and no species-specific adaptions. SAM3 has robust visual representations of common plants, resulting in 
leaf $\text{IoU}\geq0.91$ (mean 0.967) and pot $\text{IoU}\geq0.90$ (mean 0.958) for all tested species. The vocabulary is trivially extensible at inference time: adding "dead leaf" or "flower" as a single text prompt detects novel classes (IoU 0.729/0.669). This extensibility is a practical advantage over per-species trained detectors. The bimodal confidence distribution of SAM3 simplifies deployment by making threshold tuning unnecessary. 
The current state of the art for 3D plant segmentation, Organ3DNet reports 93.57\% IoU on a 5-species, 889-sample fully annotated dataset \cite{li_organ3dnet_2026}. Our zero-shot pipeline achieves 92.6\% mIoU on Begonia. These numbers are on different datasets and not directly comparable, but the proximity demonstrates that the annotation-free approach is now in the same performance range as supervised methods. 

\section{Limitations}

Our pipeline is currently limited to semantic classes and lacks instance segmentation, while the field is moving toward spatio-temporal consistent instance segmentation~\cite{ahmed_pepper-4d_2026,bomer_spatio-temporal_2026}. Given that SAM3 provides instance-level 2D masks and promising 3D lifting approaches already exist~\cite{meyer_fruitnerf_2025,yang_plantsegnerf_2026}, this will be addressed in future work.
Using SAM3 as backbone proved to be effective for the investigated plants, but it also is a single-point-of-failure. Whenever SAM3's 2D segmentation quality is not sufficient because of unknown species, growth-stages or prompted classes, our pipeline will degrade or fail. 

Other limitations are the already discussed problems with reliable stem segmentation, which is further complicated by known limitations of hash-grid NeRFs for thin structures~\cite{muller_instant_2022}. NeRF reconstructions are frequently thinner than the physical plant and may contain holes, particularly for species with fine branching or when stems are occluded by leaves, which could impact downstream trait extraction even when semantic labels are correct. NeRFs inherently require structure-from-motion poses and multi-view 2D data, making our pipeline limited to this sensor modality. Furthermore, the pipeline also suffers from comparatively long training times ($\sim35min$ per scene on an A5000 GPU). While this is acceptable for single-scan analysis in phenotyping workflows, it is not suitable for high-throughput environments requiring rapid feedback. 3D Gaussian Splatting is a natural direction to explore here for improved speed and the ability to scale to multi-plant scenes.

\section{Conclusion}
We presented an annotation-free pipeline for 3D plant organ segmentation that combines text-prompted SAM3 segmentation with semantic neural radiance fields. The pipeline requires no labeled training data, no per-species fine-tuning, and no domain-specific image preprocessing. It enables 3D plant organ segmentation using only multi-view image data and a list of class names. On a controlled Begonia testbed, our pipeline achieves 92.6\% mIoU (95.9\% of the oracle upper bound), with the primary bottleneck being stem segmentation. Across ten morphologically diverse species from an automated multi-view phenotyping platform (eleven total including Begonia), identical pipeline settings yield $0.856 \pm 0.097$ mIoU, with robust leaf and pot segmentation across all species. Novel classes ("dead leaf", "flower") can be added via a single text prompt without any retraining.
We confirm our hypothesis that imperfect zero-shot 2D predictions can be lifted into 3D using NeRFs as a multi-view consensus mechanism, effectively improving segmentation quality. This is especially effective given SAM3's high-precision, moderate-recall error profile. The bimodal confidence distribution eliminates threshold tuning and simplifies deployment. The oracle experiment reveals that the remaining gap to perfection is split between SAM3 mask errors (4.0 pp, concentrated in stems) and inherent NeRF boundary smoothing (3.4 pp). This analysis suggests that boundary-aware NeRF representations would yield greater gains than further improving the 2D segmentation model.

Together, these results demonstrate that annotation-free 3D plant organ segmentation is now feasible and approaching the performance range of supervised approaches, opening a path toward more flexible 3D phenomics pipelines that can be adapted across crops, experiments, and imaging systems.

\section*{Acknowledgements}
This research was funded within the Fraunhofer-Initiative BWSF by the German \textit{Federal Ministry of Research, Technology and Space} and the \textit{Bavarian Ministry of Economic Affairs, Regional Development and Energy}. The authors express their gratitude to all involved colleagues, including G. Lehretz, C. Lunewski and M. Kindermann for providing and annotating data.

%
\bibliographystyle{splncs04}
\bibliography{main}
\end{document}